\documentclass[10pt,twocolumn,letterpaper]{article}

\usepackage{3dv}
\usepackage{times}
\usepackage{epsfig}
\usepackage{graphicx}
\usepackage{amsmath}
\usepackage{amssymb}

\usepackage{multirow}

\usepackage[pagebackref=true,breaklinks=true,colorlinks]{hyperref}

\threedvfinalcopy 

\ifthreedvfinal\fi
\begin{document}



\title{SC6D: Symmetry-agnostic and Correspondence-free \\
6D Object Pose Estimation 
}

\author{Dingding Cai \\ Tampere University \\ {\tt\small dingding.cai@tuni.fi}
\and 
Janne Heikkilä \\ University of Oulu \\ {\tt\small janne.heikkila@oulu.fi}
\and 
Esa Rahtu \\ Tampere University \\ {\tt\small esa.rahtu@tuni.fi}
}

\maketitle

\begin{abstract}
     This paper presents an efficient symmetry-agnostic and correspondence-free framework, referred to as SC6D, for 6D object pose estimation from a single monocular RGB image. SC6D requires neither the 3D CAD model of the object nor any prior knowledge of the symmetries. The pose estimation is decomposed into three sub-tasks: a) object 3D rotation representation learning and matching; b) estimation of the 2D location of the object center; and c) scale-invariant distance estimation (the translation along the z-axis) via classification. 
     SC6D is evaluated on three benchmark datasets, T-LESS, YCB-V, and ITODD, and results in state-of-the-art performance on the T-LESS dataset. 
     Moreover, SC6D is computationally much more efficient than the previous state-of-the-art method SurfEmb. The implementation and pre-trained models are publicly available at \href{https://github.com/dingdingcai/SC6D-pose}{https://github.com/dingdingcai/SC6D-pose}.

\end{abstract}

\section{Introduction}


Estimating the 6D pose of an object from an RGB image is one of the long-standing problems in computer vision. In this task, the objective is to infer the geometric transformation, \ie, a 3D rotation and a 3D translation, from the object coordinate system to the camera coordinate system \cite{hinterstoisser2011multimodal, hodan2020bop}. Object 6D pose estimation is one of the core components in many applications, such as robotic manipulation \cite{collet2011moped,tremblay2018deep}, augmented reality \cite{marchand2015pose}, and autonomous driving \cite{wu20196d}.

Recent learning-based approaches \cite{he2021ffb6d, di2021so, shugurov2021dpodv2} have utilized one-to-one 2D-3D correspondences to obtain excellent pose estimation results, given ambiguity-free image of the object. However, the performance of these correspondence based methods deteriorates quickly if the object exhibits visual ambiguities due to, for example, symmetries, challenging or missing textures, and occlusions.

\begin{figure}[t]
\centerline{\includegraphics[width=1.0\linewidth]{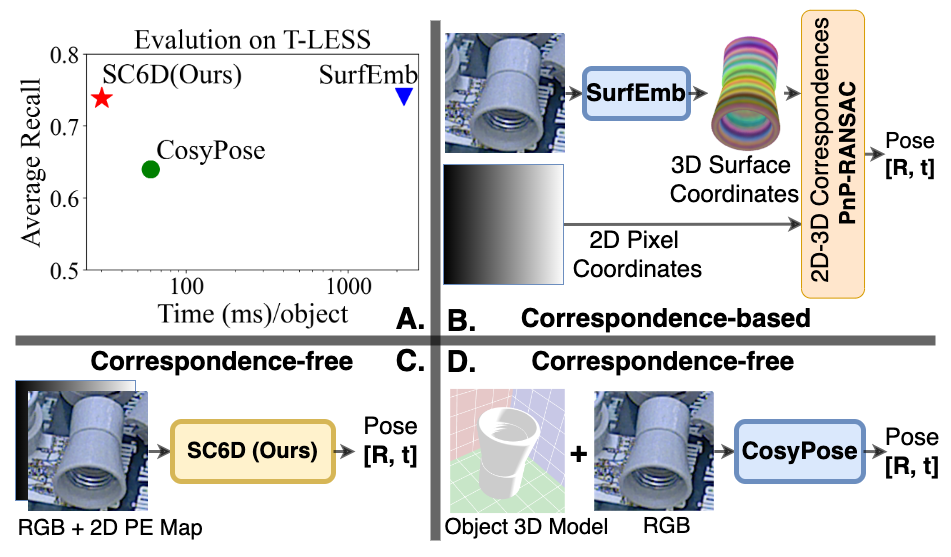}}
\caption{
Comparison between the proposed SC6D, the correspondence-based SurfEmb \cite{haugaard2021surfemb},  and the correspondence-free CosyPose \cite{labbe2020cosypose}. Unlike SurfEmb, SC6D can directly infer the 6D pose to ensure the efficiency (much faster than SurfEmb). Compared to CosyPose, SC6D does not require the object 3D model or prior knowledge of the object symmetries.
} 
\label{fig:intro}
\end{figure}

Instead of using one-to-one correspondences, Haugaard and Buch proposed a method called SurfEmb \cite{haugaard2021surfemb} which handled the visual ambiguities implicitly by learning one-to-many 2D-3D correspondence distribution. SurfEmb resulted in the state-of-the-art performance on the challenging T-LESS dataset \cite{hodan2017tless}, which consists of texture-less and symmetric industrial objects with heavy occlusion. However, SurfEmb is computationally heavy due to the time-consuming PnP-RANSAC \cite{fischler1981random} procedure (see Fig.\ref{fig:intro}).



Alternatively, Labb{\'e} \emph{et al.}~proposed a method called CosyPose \cite{labbe2020cosypose}, where the 6D pose is inferred directly without explicitly established correspondences. CosyPose provided competitive accuracy, but with substantially reduced computational load. 
However, the method requires the object 3D CAD model and prior knowledge of the object ambiguities, such as symmetries, to choose appropriate ambiguity-aware training loss. At the inference time, the object CAD model is also needed for rendering a synthetic image that is concatenated with the input image to refine the 6D pose estimate. 
These requirements may be inconvenient for many use cases.

The main challenge for training an ambiguity-agnostic 6D pose estimator comes from the fact that 
objects with symmetries have the same appearance from multiple different poses, and the mapping between the image and the pose is no longer bijective without prior information.
Recently, Murphy \etal proposed a method called Implicit-PDF \cite{murphy2021implicit} to represent the distributions on the rotation manifold. Their method concatenates the object representation vector with each 3D rotation sampled from SO(3) space and feeds the combined vector to an MLP network to compute a log probability. Implicit-PDF can estimate the 3D rotation distribution of the object from an RGB image without knowing any prior information about ambiguities. However, their approach concatenates millions of rotation samples to each object representation vector and feeds them to the MLP to compute the distribution online, which leads to inferior computing efficiency.

This paper presents a correspondence-free approach, called SC6D, for symmetry-agnostic 6D object pose estimation using SO(3) embedding. Different from Implicit-PDF, SC6D aims to learn a latent representation for each 3D rotation sample in SO(3) space and associate the rotation with the object visual representation in the latent space based on the cosine similarity. Moreover, SC6D can simultaneously estimate the 3D translation from the object coordinate system to the camera reference frame.

To speed up the inference, SC6D constructs a 3D rotation representation library for each object in an offline manner. The object 3D rotation is predicted by matching the object representation with the rotation embeddings stored in the library. Meanwhile, the 3D translation is recovered by regressing the offset from the object projection center to the object bounding box center and classifying the object translation along the z-axis into the pre-defined distance bins.

We evaluate SC6D on three benchmark datasets, T-LESS \cite{hodan2017tless}, YCB-V \cite{xiang2018posecnn}, and ITODD \cite{drost2017introducing}, and compare it with several recent baselines. SC6D achieves state-of-the-art performance (78.0\% average recall) on the challenging T-LESS dataset and is comparable to the baselines on the YCB-V and ITODD datasets.


In summary, the main contributions are: 1) a method for learning the latent 3D rotation representation to implicitly handle the object symmetries; 2) an efficient and effective correspondence-free framework for symmetry-agnostic object 6D pose estimation; and 3) a scale-aware and classification-based (z-axis) translation estimation solution.


\section{Related Work}
Object 6D pose estimation is a wide field of research with numerous previous works. This section focuses on recent learning based monocular RGB(-D) based methods, which are most relevant to the proposed approach. 

\subsection{Correspondence-based 6D pose estimation}
The correspondence-based methods  \cite{brachmann2014learning,tekin2018real,pavlakos20176,peng2020pvnet,rad2017bb8,Park_2019_ICCV,zakharov2019dpod,hodan2020epos,Wang_2021_GDRN, song2020hybridpose} are the dominating approach for the 6D object pose estimation. These methods attempt to establish 2D-3D (3D-3D) correspondences between the 2D RGB (Depth) image pixel coordinates and the 3D coordinates on the object 3D model surface, followed by solving for the 6D pose using a variant of Perspective-n-Point(PnP) \cite{lepetit2009epnp} algorithm (or a least-squares fitting \cite{arun1987least} algorithm for 3D-3D correspondences). Two types of strategies are mainly applied among these methods, \ie, sparse correspondence-based and dense correspondence-based.
\paragraph{Sparse correspondence-based methods.} These methods \cite{rad2017bb8,peng2019pvnet,song2020hybridpose,peng2020pvnet,tekin2018real,he2021ffb6d,pvn3d} are also called keypoint-based approaches which detect the predefined keypoints (known 3D coordinates on the object 3D model) from the input data. BB8 \cite{rad2017bb8} was proposed by Rad \etal to detect the 2D projection coordinates of the object 3D bounding box corners (keypoints) in the RGB image. The object 6D pose is then recovered from the established spare 2D-3D correspondences using PnP \cite{lepetit2009epnp}. Another approach \cite{tekin2018real} called YOLO6D also attempts to detect the object 3D bounding box corner projections but with a more efficient detector \cite{yolov2} and thus is able to run in real-time. However, both BB8 and YOLO6D are incapable of handling the object occlusion. Peng \etal proposed an occlusion-robust approach PVNet \cite{peng2019pvnet}, which predicts the pixel-wise voting vectors to localize the keypoints defined based on the object 3D model instead of on the 3D bounding box. The following works \cite{pvn3d,he2021ffb6d} further extended the vote-based idea to build the spare 3D-3D correspondences from the depth data and have achieved a saturated performance ($>=$99.4\%) on the LineMOD \cite{hinterstoisser2011multimodal} dataset by solving the pose with the least-square fitting algorithm \cite{arun1987least}. Nevertheless, these keypoint-based methods quickly fail to handle the symmetric objects, especially when the symmetries are unknown.
\paragraph{Dense correspondence-based methods.} Many recent works \cite{Park_2019_ICCV,li2019cdpn,hodan2020epos,zakharov2019dpod,shugurov2021dpodv2,Wang_2021_GDRN,haugaard2021surfemb} focus on learning the dense pixel-wise correspondences to improve the accuracy of the estimated 6D pose. Park \etal \cite{Park_2019_ICCV} propose a regression-based method, Pix2Pose, for predicting the pixel-wise 2D-3D correspondences from a single RGB image. Knowing the dense correspondences, Pix2Pose computes the pose using the EPnP \cite{lepetit2009epnp} with the RANSAC \cite{fischler1981random} algorithm.
CDPN \cite{li2019cdpn} is also regression-based method for dense 2D-3D correspondence prediction. Instead of using PnP for recovering both 3D rotation and 3D translation, CDPN directly regresses the 3D translation parameters and solves PnP for 3D rotation based on the predicted correspondences. In contrast, DPOD \cite{zakharov2019dpod} presented a classification-based correspondence estimation framework, which turned be a more effective strategy than regression. However, these approaches are incapable of handling various object symmetries, especially when the symmetries are unknown. To better handle the object ambiguities, EPOS \cite{hodan2020epos} proposes to learn a  probability distribution of the 2D-3D correspondences over the object surface fragments. EPOS first categorizes an object pixel into a predefined object surface fragment and then regresses the coordinate offset within the fragment it belongs to. An efficient GC-RANSAC \cite{barath2018graph} algorithm is employed to estimate the pose from the correspondences. Another latest distribution-based method SurfEmb \cite{bay2006surf} is presented by Haugaard and Buch to learn a dense and continuous correspondence distribution without knowing any prior information about object visual ambiguities. SurfEmb associates the 2D pixel coordinates of the object with the 3D surface coordinates on the object CAD model in the feature embedding space to establish the pixel-wise one-to-many correspondences, based on which the object pose is later estimated and refined using PnP-RANSAC \cite{fischler1981random}. State-of-the-art results are achieved by SurfEmb on several benchmark datasets in the BOP challenge \cite{hodan2020bop} but with a significant runtime cost.


\subsection{Correspondence-free 6D pose estimation}
There are also many attempts \cite{xiang2018posecnn,kehl2017ssd,wang2019densefusion,li2018deepim,labbe2020cosypose,Sundermeyer_2018_ECCV,sundermeyer2020multi,cai2022ove6d,wohlhart2015learning,zakharov20173d} to use the neural networks to estimate the object pose without manually solving the PnP problem based on the intermediate 2D-3D correspondence representation.

\paragraph{Regression-based methods.} These learning-based methods \cite{xiang2018posecnn,wang2019densefusion,Wang_2021_GDRN,di2021so,li2018deepim,labbe2020cosypose} regress the 6D pose parameters based on the success of deep neural networks. Xiang \etal \cite{xiang2018posecnn} proposed an early well-known work, called PoseCNN, for object 6D pose estimation from RGB images by decoupling the task into several regression-based sub-tasks. In addition, GDR-Net \cite{Wang_2021_GDRN} and SO-Pose \cite{di2021so} benefit from the geometry-guided information and have achieved impressive performance on the monocular RGB image-based pose estimation task by using regression. When the object 3D model is available, DeepIM \cite{li2018deepim} proposes to regress a relative pose between the input image and the rendered synthetic object image to refine the 6D pose. The follow-up work CosyPose \cite{labbe2020cosypose} further improved the pose accuracy by using a new continuous rotation representation \cite{zhou2019continuity} and achieved impressive performance on several benchmark datasets from BOP \cite{hodan2020bop}. The main advantage of the regression-based method is its high efficiency at the inference time. The object translation parameters are estimated in a regression-based (and classification-based) manner in this work.

\paragraph{Template-based methods.} Benefiting from the powerful representation learning capability of modern neural works, these approaches \cite{Sundermeyer_2018_ECCV,sundermeyer2020multi,sundermeyer2020augmented,cai2022ove6d,wohlhart2015learning} attempt to learn the object pose-aware representations from the input data using the backbone networks, like ResNet \cite{he2016deep}. This type of approach aims at constructing a pose-aware template library for the target object. Each template is a representation vector
extracted from the image of the object associated with a specific pose. At the inference time, the object pose can be recovered by matching the representation extracted from the input image with templates in the library. Sundermeyer \etal proposed an augmented auto-encoder (AAE) \cite{Sundermeyer_2018_ECCV} for 
object pose-aware representation encoding from the RGB images. A separate AAE model was trained per object, which is a cumbersome task for multiple objects. Their follow-up work MP-Encoder \cite{sundermeyer2020multi} is thus proposed to mitigate the problem, and a single MP-Encoder accounts for multiple objects in a dataset. However, a large number of templates ($>$90,000) are needed for both AAE and MP-Encoder to cover all possible views of the object. In contrast, Cai \etal \cite{cai2022ove6d} proposed a depth-based approach called OVE6D for encoding the object viewpoint only, which can significant reduce the number of templates (\eg 4000) by ignoring the object in-plane rotation. These template-based methods can inherently handle object symmetries due to the similarity-based matching strategy. To this end, we attempt to learn an SO(3) encoder for extracting the rotation representation and constructing the 3D rotation templates. Different from the above methods, the rotation representation vector is extracted from the corresponding 3D rotation matrix (instead of from RGB(D) images) by the SO(3) encoder.

\section{Method}

\subsection{Overview}

\begin{figure}[t]\centerline{\includegraphics[width=0.85\linewidth]{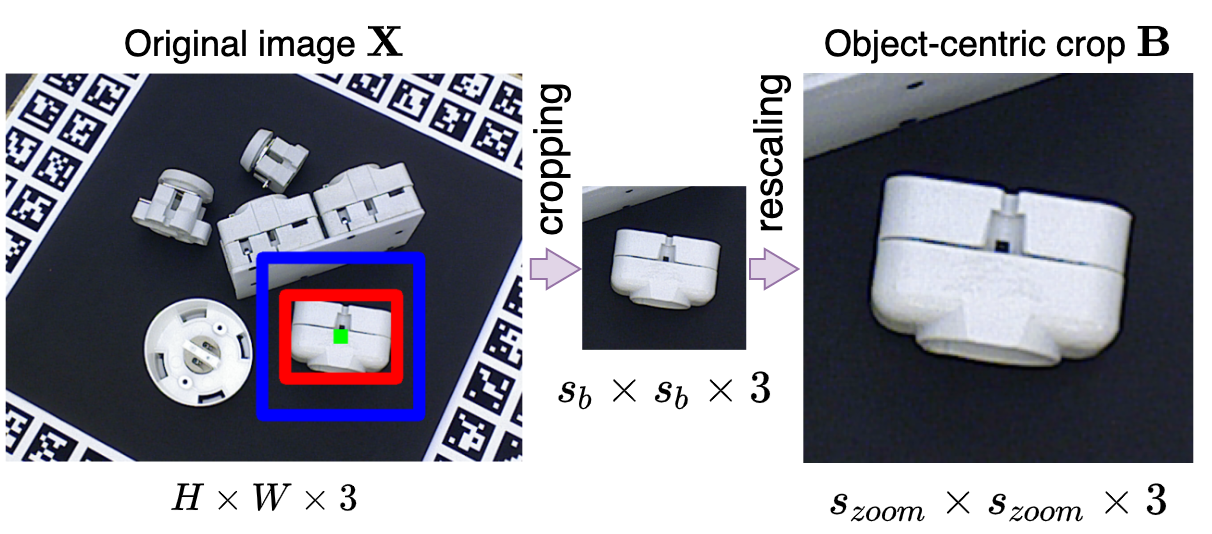}}
\caption{
Illustration of cropping the object-centric region $\mathbf{B}$ from the original RGB image $\mathbf{X}$. The red inner bounding box (centered at $(b_x, b_y$)) is predicted by off-the-shelf detectors such as YOLO \cite{redmon2018yolov3} or Faster-RCNN \cite{ren2015faster}. Following \cite{Wang_2021_GDRN}, the object is cropped using an enlarged bounding box (outer blue box) with the size $s_{b}=f_p \cdot  max(b^{'}_w, b^{'}_h)$, where $(b^{'}_w, b^{'}_h)$ is the predicted bounding box size and $f_p$ is the padding factor ($f_p=1.5$ in our experiments), and then rescaled to the desired size $s_{zoom}$. By doing so, we keep the object aspect unchanged.
}\label{fig:cropping}
\end{figure}

The goal is to estimate the 6D pose $\mathbf{P}=[\mathbf{R} | \mathbf{t}]$ of an object, where $\mathbf{R}\in SO(3)$ is the 3D orientation and $\mathbf{t}\in R^3$ is the 3D location, from an RGB image $\mathbf{X} \in R^{H\times W \times 3}$.
Instead of predicting the object egocentric orientation $\mathbf{R}$ (\ie orientation \textit{w.r.t} the camera) from an RGB image, we estimate the object allocentric orientation $\mathbf{R}_{allo}$ \cite{kundu20183d}, \ie orientation \textit{w.r.t} the object, because objects with similar allocentric orientation have similar appearance, which is important for learning the object representation. The rotation is estimated from the cropped (and rescaled) object-centric RGB image region $\mathbf{B}\in R^{s_{zoom}\times s_{zoom} \times 3}$ (see Fig.\ref{fig:cropping}) based on the object bounding box predicted by off-the-shelf detectors such as YOLO \cite{redmon2018yolov3} or Faster-RCNN \cite{ren2015faster}. 
The egocentric orientation can be recovered by $\mathbf{R}=\mathbf{R}_c \mathbf{R}_{allo}$, where $\mathbf{R}_c$ is the 3D rotation between the ray through the object center and the camera optical axis.

\begin{figure*}[t]\centerline{\includegraphics[width=0.9\linewidth]{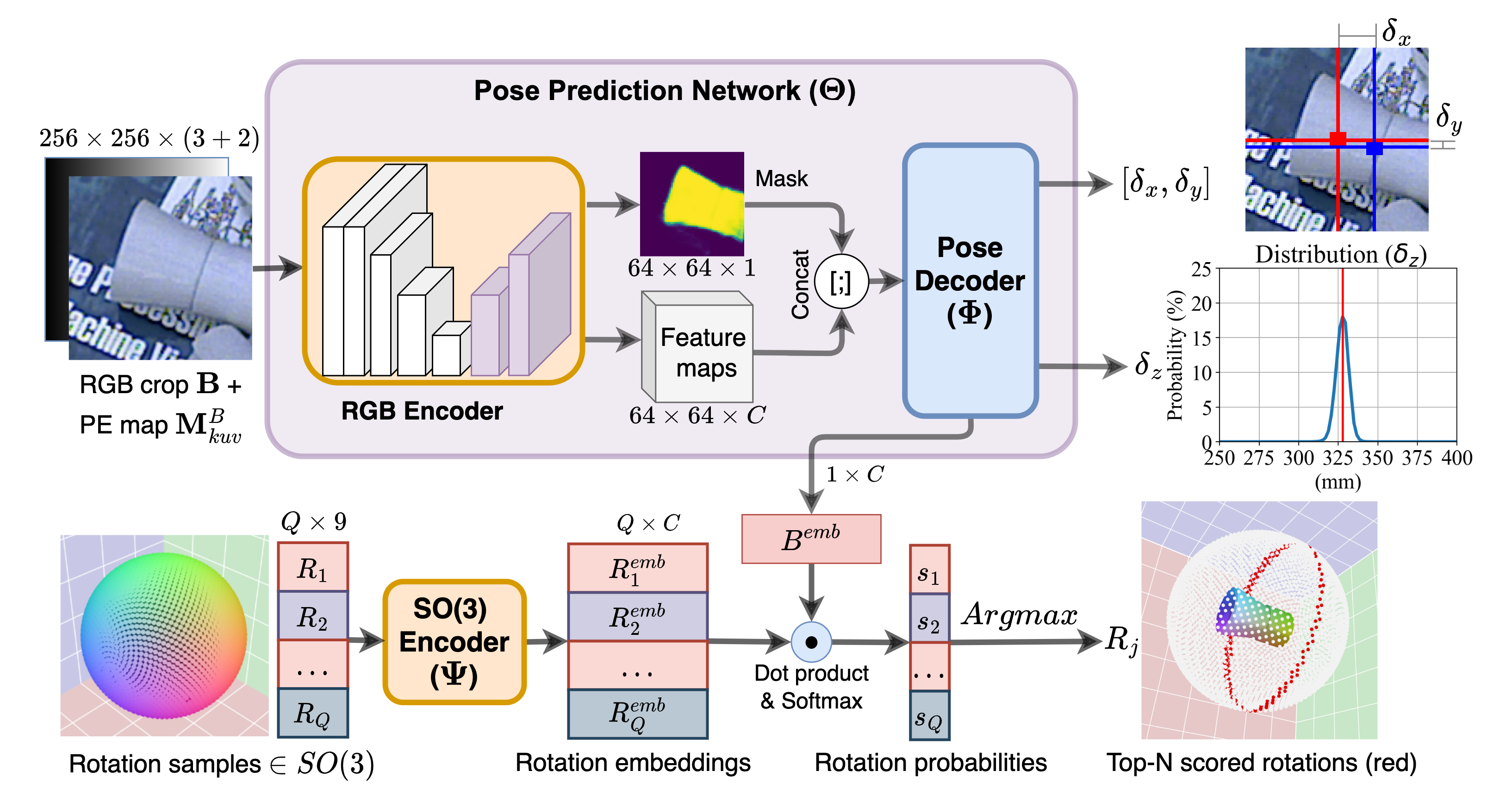}}
\caption{\textbf{The overview of SC6D}. The RGB Encoder takes an RGB crop $\mathbf{B}$ concatenated with a 2D positional encoding (PE) map $\mathbf{M}^B_{kuv}$ as input and outputs a feature map accompanying with a segmentation mask. The feature map is concatenated with the object segmentation mask and is fed into the Pose Decoder $\mathbf{\Phi}$. The Pose Decoder outputs an object projection offset $[\delta_x, \delta_y]$,  a location distribution $\delta_z$ along the z-axis, and a representation vector $B^{emb}$. The SO(3) Encoder $\mathbf{\Psi}$ takes the 3D rotation matrices $\in SO(3)$ (sampled from SO(3) space) as input and outputs the rotation embeddings for the object 3D orientation estimation. The rotation with the highest probability is directly selected as the object orientation estimation. Note that the top-N rotations with the highest probabilities are shown as red points on the sphere for visualisation only.
} 
\label{fig:overview}
\end{figure*}

SC6D estimates the object 3D orientation distribution and the object 3D location from an RGB image crop without using the object 3D model.  The method is comprised of three modules: 1) an RGB Encoder, 2) a Pose Decoder, and 3) an SO(3) Encoder. The overall architecture of SC6D is illustrated in Figure \ref{fig:overview}. The following subsections provide a detailed description of the model components and the loss functions used in model training.  



\subsection{Learning 3D Orientation Distribution}
It is non-trivial to estimate the 3D orientation for symmetric objects even if the prior knowledge of object symmetry is known. In this work, we propose to learn an SO(3) encoder for 3D orientation estimation to avoid explicitly handling object symmetries. 

Given the object-centric image crop $\mathbf{B}$, we aim to learn the object 3D orientation distribution based on its visual representation, \ie,
\begin{equation}
\label{eq:R_dist}
p(\mathbf{R}_{allo}|\mathbf{B})=\frac{ p(\mathbf{R}_{allo},\mathbf{B}) } 
{ \int_{R\in SO(3)} p(R, \mathbf{B}) \, dR }
\end{equation}
We approximate Equation \ref{eq:R_dist} by 
\begin{equation}
\label{eq:R_app}
p(\mathbf{R}_{allo}|\mathbf{B}) \approx \frac{ 
exp(\mathbf{\Theta}(\mathbf{B}) \cdot \mathbf{\Psi}(\mathbf{R}_{allo}) )
} {
\sum^Q_i exp(\mathbf{\Theta}(\mathbf{B}) \cdot \mathbf{\Psi}(\mathbf{R}_i))
},
\end{equation}
where $\mathbf{R}_{allo}\in R^{3\times 3}$ is the object allocentric orientation, $\{\mathbf{R}_i \}^Q_{i=1} \in SO(3)$ are the rotation samples, $\mathbf{\Theta}$ represents the object pose prediction network, and $\mathbf{\Psi}$ is the SO(3) encoder network, as shown in Figure \ref{fig:overview}. 

\subsection{Scale-Invariant Location Estimation}


We denote $\mathbf{\bar{P}}^B \in R^3$ and $\mathbf{\bar{P}}^X \in R^3$ as the projection coordinates \footnote{The 2D coordinates are represented in the homogeneous form.} of the object 3D origin point in the crop $\mathbf{B}$ and the original image $\mathbf{X}$, respectively. To associate $\mathbf{\bar{P}}^B$ with $\mathbf{\bar{P}}^X$, we formulate the equation as 
\begin{equation}
\begin{split}
\mathbf{\bar{P}}^{B} = \mathbf{T}^{B}_{X}\mathbf{\bar{P}}^{X} = \begin{bmatrix} s_{zoom}  \delta_x \\ s_{zoom}\delta_y  \\ 1 \end{bmatrix}, 
\end{split}
\end{equation}
where $\mathbf{T}^{B}_{X}\in R^{3\times3}$ represents the transformation from the original image coordinate system $\mathbf{X}$ to the object-centric coordinate system $\mathbf{B}$ of the image crop \footnote{For simplicity, we abuse the notations for RGB images and the corresponding coordinate systems.}. Similar to the Scale-Invariant Translation Estimation (SITE) proposed in  \cite{Wang_2021_GDRN}, $(\delta_x, \delta_y)$ represents the proportional offsets from the projection of the object origin to the crop ($\mathbf{B}$) center $(s_{zoom}/2, s_{zoom}/2)$ and is invariant to the coordinate transformation $\mathbf{T}^B_X$. In this work, the offset $(\delta_x, \delta_y)$ is directly regressed from the crop $\mathbf{B}$ (see the top-right picture in Fig. \ref{fig:overview}). 

The transformation $\mathbf{T}^{B}_{X}$ can be viewed as a 2D translation and scaling operation, 
\begin{equation}
\begin{split}
\mathbf{T}^B_X
&= \begin{bmatrix} 
        r & 0 & - rb_x \\
        0 & r & - rb_y \\
        0 & 0 &     1     \\
   \end{bmatrix},
\end{split}
\end{equation}
where $r = s_{zoom} / s_b$ is the scaling factor of the object-centric crop $\mathbf{B}$ from the original object bounding box scale $s_b$ to the target scale $s_{zoom}$, and $(b_x, b_y)$ is the object bounding box center in the original image. 
Thereby, we can obtain a new calibrated (virtual) camera intrinsic matrix $\mathbf{K}_B$ associated with each object-centric crop $\mathbf{B}$,
\begin{equation}
\begin{split}
\mathbf{K}_B &= \mathbf{T^{B}_{X}}\mathbf{K}_X
= \begin{bmatrix} 
rf_x &    0  & r(c_x - b_x) \\
   0  & rf_y & r(c_y - b_y) \\
   0  &    0  &         1           \\
\end{bmatrix},
\end{split}
\end{equation}
where $\mathbf{K}_X$ is the original camera intrinsic matrix, $(f_x, f_y)$ and $(c_x, c_y)$ are the original camera focal length and principal point. 
Thus, the 2D positional encoding (PE) map 
for the input crop $\mathbf{B}$ can be obtained by $\mathbf{M}^B_{kuv}=\mathbf{K}_B^{-1}\mathbf{G}_{uv}$, where $\mathbf{G}_{uv}\in R^{s_{zoom}\times s_{zoom}\times 3}$ is the 2D grid of (homogeneous) coordinates of the crop $\mathbf{B}$. 

We transform (project) the object origin point to the image coordinate system $\mathbf{B}$ by leveraging the new intrinsic matrix $\mathbf{K}_B$,
\ie, $\mathbf{K}_B(\mathbf{R} P_o + \mathbf{t})= t_z \bar{\mathbf{P}}^B$,
where $P_o$ is the object origin point ($[0,0,0]^T$) in the object coordinate system, $t_z$ is the object translation (location) along the z-axis. Intuitively, $t_z$ is directly associated with the object scale in RGB images. However, the object scale (appearance) is changed when we rescale the object crop from the original scale $s_b$ to the target scale $s_{zoom}$. To this end, we estimate the scale-invariant parameter $\delta_z=t_z / r$, where $r$ is the rescaling factor, for the object z-axis translation as in  \cite{Wang_2021_GDRN} from the rescaled object-centric crop $\mathbf{B}$. Thus, the 3D translation from the object coordinate system to the camera coordinate system can be recovered as $\mathbf{t} = r \delta_z \mathbf{K}_B^{-1} \bar{\mathbf{P}}^B$. 

Instead of directly regressing the $\delta_z$, we formulate the z-axis translation estimation as a classification task that has been proved to be more effective and successful in depth estimation  \cite{fu2018deep}. To this end, we uniformly discretize the object location along the z-axis into $K$ bins, \ie $d_i = d_l + (d_u - d_l)*i/K$
, where $i$ is the bin index ranging from $0$ to $K-1$, $d_l$ and $d_u$ are the lower bound and the upper bound of $\delta_z$ ($\delta_z \in (d_u, d_l)$) in the training data, respectively.
Alternatively, we can also calculate the expectation and obtain a continuous $\delta_z$ by weighting the class labels, \ie, $\delta_z=\sum^{K-1}_{i=0}p_id_i$, where $p_i$ is the classification probability for the class label $d_i$ (see the middle right picture in Fig.\ref{fig:overview}).

\subsection{Recovering the Egocentric Orientation}
In this work, the allocentric orientation ($\mathbf{R}_{allo}$ $\textit{w.r.t}$ the object) is estimated from the object-centric RGB crop $\mathbf{B}$. However, the 6D object pose is represented conventionally in an egocentric form ($\mathbf{R}$ \textit{w.r.t} the camera). Therefore, we need to estimate the rotation matrix $\mathbf{R}_c$ to recover the egocentric orientation \cite{kundu20183d}, \ie, $\mathbf{R}=\mathbf{R}_c \mathbf{R}_{allo}$.

The rotation matrix $\mathbf{R}_c$ can be obtained by estimating the 3D rotation from the ray $\mathbf{o}_{ray}=\mathbf{K}_B^{-1} \mathbf{\bar{P}}^{B}$ through the object origin to the camera optical axis $\mathbf{c}_{ray}=[0, 0, 1]^T$ in the original image $\mathbf{X}$,
\begin{equation} \label{eq:allo}
\begin{split}
\mathbf{R}_c = \mathbf{\textit{I}} + R_r + \frac{R^2_r}{1 + \mathbf{c}_{ray} \cdot \mathbf{\bar{o}}_{ray} }
\end{split}
\end{equation}
where  $\mathbf{\mathit{I}} \in R^{3\times 3}$ is the identity matrix,
$\mathbf{\bar{o}}_{ray}=\mathbf{o}_{ray}/|\mathbf{o}_{ray}|$
is the normalized unit vector, and $R_r$ is the skew-symmetric matrix of the vector $\mathbf{r} = \mathbf{c}_{ray} \times \mathbf{o}_{ray}$.

\subsection{Loss functions}
We adapt the InfoNCE loss \cite{van2018representation} to learn the 3D orientation representation and distribution:
\begin{equation}
\begin{split}
\mathcal{L}_R &= -log (p(\mathbf{R}_{allo}|\mathbf{B} )) \\
&= -log \frac{exp( B^{emb} \cdot R^{emb}_{allo} / \tau )} {\sum^Q_i exp( B^{emb} \cdot R^{emb}_i / \tau)},
\end{split}
\end{equation}
where $ B^{emb}$ is the normalized representation vector extracted from the object-centric crop $\mathbf{B}$, $R^{emb}_{allo}$ is the normalized embedding representation of the ground truth allocentric orientation $\mathbf{R}_{allo}$, $\{R^{emb}_i\}^Q_{i=1}$ are the normalized embedding vectors for the orientations sampled from SO(3) space, and $\tau$ is the temperature parameter.

We employ the L1 loss for training the object projection offset $(\delta_x, \delta_y)$ prediction task and the focal loss  \cite{lin2017focal} for training the object z-axis location ($\delta_z$) classification task. Thereby, 
\begin{equation}
\begin{split}
\begin{cases}
\mathcal{L}_{xy} &= ||\delta_x - \hat{\delta}_x||_1 + ||\delta_y - \hat{\delta}_y||_1 \\
\mathcal{L}_z &= -\alpha (1-\mathbf{p}_z \cdot \mathbf{\hat{p}}_z)^{\gamma}log(\mathbf{p}_z \cdot \mathbf{\hat{p}}_z)
\end{cases},
\end{split}
\end{equation}
where $(\hat{\delta}_x, \hat{\delta}_y)$ are the ground truth offsets, $\mathbf{p}_z \in R^K$ is the network output probability,  $\mathbf{\hat{p}}_z \in R^K$ is the one-hot vector for the ground truth label $\hat{\delta}_z$, $\alpha$ and $\gamma$ are the hyperparameters for the focal loss.

In addition, we use the average binary cross entropy loss $\mathcal{L}_M$ for training to predict the object segmentation mask. The total loss is hence written as 
\begin{equation}
\begin{split}
\mathcal{L} = \lambda_R \mathcal{L}_R + \lambda_M \mathcal{L}_M +  \lambda_{xy} \mathcal{L}_{xy}  +  \lambda_z \mathcal{L}_z ,
\end{split}
\end{equation}
where $\lambda_{\{R, M, xy, z\}}$ are the loss balancing parameters. 

\section{Experiments}

We conduct extensive experiments on three datasets, \ie, T-LESS \cite{hodan2017tless}, YCB-V \cite{xiang2018posecnn}, and ITODD \cite{drost2017introducing}, to demonstrate the effectiveness of SC6D. In addition, the ablation studies on T-LESS are carried out to verify the effectiveness of each individual component.


\paragraph{Datasets.} 
The T-LESS dataset  \cite{hodan2017tless} provides 30 texture-less industrial objects accompanying with the object 3D models. Each image contains multiple texture-less objects with heavy occlusion and various symmetries. YCB-V contains 21 objects with richer textures and fewer symmetries compared to T-LESS. ITODD includes 28 realistic industrial objects which are captured by gray-scale images. The evaluations are conducted on the testing subsets following the BOP challenge protocol \cite{hodan2020bop}, and the prediction results are submitted to the public BOP challenge platform \footnote{https://bop.felk.cvut.cz/challenges} for evaluation to ensure a fair comparison to other methods. 

\paragraph{Architecture details.} The entire architecture consists of an RGB encoder, a pose decoder, and an SO(3) encoder. The RGB encoder is an asymmetric UNet \cite{ronneberger2015u} comprised of a pretrained ResNet34 \cite{he2016deep} backbone with a separate decoding head per object. The pose decoder is shared by all objects and is comprised of the shared Conv2D layers followed by three parallel MLP headers with the output dimensions $C_{xy}=2$ for the projection offset, $C_z=K$ for the z-axis location classification and $C_R=32$ for the object representation vector. The SO(3) encoder is a lightweight MLP network and has three fully-connected layers including an input layer ($C_{i}=256$), a hidden layer ($C_{h}=256$), and a separate output layer ($C_{o}=32$) per object. More architecture details are shown in the supplementary materials.

\paragraph{Implementation details.} We implement SC6D using Pytorch \cite{paszke2019pytorch} framework and train the model using AdamW solver \cite{loshchilov2017decoupled} with the cosine annealing learning rate starting from $5 \times 10^{-4}$ to $1 \times 10^{-5}$ and weight decay $1 \times 10^{-4}$ for 75 epochs on 16 Nvidia GPUs. In all our experiments, we set the hyper-parameters $K=1000$, $\tau=0.1$, $\lambda_R=1.0$, $\lambda_M=1.0$, $\lambda_{xy}=10.0$, $\lambda_z=1.0$, $\alpha=0.5$, and $\gamma=2$ to ensure comparable magnitudes among all training loss terms. Moreover, we set $Q=5000$ rotation samples for training the SO(3) encoder and use $Q=480,000$ rotation samples (uniform sampling) for constructing the rotation representation library. We train a single model for each dataset using the synthetic Physically-Based Rendering (PBR) images provided by the BOP challenge \cite{hodan2020bop}. Note that we also apply strong image augmentation strategies as in \cite{Wang_2021_GDRN} during training.
Since both T-LESS and YCB-V provide real training images, we also fine-tune the model for additional 30 epochs on a mixture of synthetic and real training images. 
During inference, we follow SurfEmb \cite{haugaard2021surfemb} and utilize the predicted object bounding boxes provided by CosyPose \cite{labbe2020cosypose} for a fair comparison \footnote{The default detection results also used for BOP Challenge 2022.}. We also report the results using the simple test-time augmentation (similar to SurfEmb), \ie, rotating the input image by $0^\circ$, $90^\circ$, $180^\circ$, and $270^\circ$ to obtain the predictions, then rotating the predictions back. In this case, we select the rotation sample with the highest probability along with the mean of the 3D location prediction as the object 6D pose estimate.

\paragraph{Evaluation metrics.} We follow the BOP challenge  \cite{hodan2020bop} and adopt the standard 6D pose estimation metric to evaluate the pose accuracy based on three pose errors, \ie, Visible Surface Distance (VSD), Maximum Symmetry-aware Surface Distance (MSSD), and Maximum Symmetry-aware Projection Distance (MSPD). We compute an average recall for each of the pose errors, \ie, AR\textsubscript{VSD}, AR\textsubscript{MSSD}, and AR\textsubscript{MSPD}, based on the standard error thresholds. We average these three recalls to obtain an overall (mean) Average Recall (AR). The detailed definitions of these metrics can be found in the paper \cite{hodan2020bop}.

\begin{table*}[!ht]
\centering
\small
\renewcommand{\arraystretch}{1.2}
\begin{tabular}{c | c | c | c | c c | c c c | c }
\hline
\shortstack{ ~ \\ Category \\ ~ } & \shortstack{ ~ \\ Method \\ ~ } & \shortstack{ ~ \\ Object \\3D model}  & \shortstack{ ~ \\ SymPrior \\ ~ }  & \shortstack{ ~ \\ Synt \\ ~} & \shortstack{  Real \\ ~ } & \shortstack{ ~ \\ T-LESS\\ ~ } & \shortstack{ ~ \\ ITODD \\ ~ } & \shortstack{ ~ \\  YCB-V \\ ~ } & \shortstack{ ~ \\ Avg (AR) \\ ~ }\\ 
\hline
\hline
\multirow{8}{*}{\shortstack{ ~\\ Correspondence \\ based methods \\ ~}}
& CDPNv2 \cite{li2019cdpn}                 & \checkmark &  & \checkmark &  & 0.407 & 0.102 & 0.390 & 0.300 \\
& EPOS \cite{hodan2020epos}                & \checkmark &  & \checkmark &  & 0.467 & 0.186 & 0.499 & 0.384 \\
& ZebraPose \cite{su2022zebrapose}         & \checkmark &  & \checkmark &  & 0.603 &   -   &   -   &   -   \\
& DPODv2 \cite{shugurov2021dpodv2}         & \checkmark & \checkmark  & \checkmark & & 0.636 & - & - & -   \\
& SurfEmb \cite{haugaard2021surfemb}       & \checkmark &  & \checkmark &  & 0.741 & 0.387 & 0.653 & 0.594 \\
\cline{2-10}
& CDPNv2 \cite{li2019cdpn}                & \checkmark  & & \checkmark & \checkmark & 0.478  & 0.102 & 0.532 & 0.371 \\
& EPOS \cite{hodan2020epos}               & \checkmark  & & \checkmark & \checkmark & 0.476  & 0.186 & 0.696 & 0.453 \\
& SurfEmb \cite{haugaard2021surfemb}      & \checkmark  & & \checkmark & \checkmark & 0.770  & \textbf{0.387} & 0.718 & \textbf{0.625} \\
\hline\hline
\multirow{8}{*}{\shortstack{Correspondence \\ free methods}}
& CosyPose \cite{labbe2020cosypose}{\dag} & \checkmark & \checkmark & \checkmark & & 0.520 & 0.131 & 0.334 & 0.328 \\
& CosyPose \cite{labbe2020cosypose} & \checkmark & \checkmark & \checkmark & & 0.640 & 0.216 & 0.574 & 0.477 \\
& SC6D (Ours){*}  &            &     & \checkmark &  & 0.729  & 0.295 & 0.594 & 0.539 \\
& SC6D (Ours)     &            &     & \checkmark &  & 0.739  & 0.303 & 0.610 & 0.551 \\ 
\cline{2-10}
& CosyPose \cite{labbe2020cosypose}{\dag} & \checkmark & \checkmark & \checkmark & \checkmark & 0.616 & 0.131 & 0.655 & 0.467 \\
& CosyPose \cite{labbe2020cosypose} & \checkmark & \checkmark & \checkmark & \checkmark & 0.728  & 0.216 & \textbf{0.821} & 0.588 \\
& SC6D (Ours){*}   &            &     & \checkmark & \checkmark & \underline{0.771} & 0.295 & 0.781 & 0.616 \\
& SC6D (Ours)      &            &     & \checkmark & \checkmark & \textbf{0.780}  & \underline{0.303} & \underline{0.788} & \underline{0.624} \\ 
\hline
\end{tabular}
\vspace{2px}
\caption{Evaluations on \textbf{T-LESS}, \textbf{ITODD}, and \textbf{YCB-V} in terms of the Average Recall (AR) \cite{hodan2020bop}. 
SymPrior: the prior knowledge of object symmetries. Synt: Synthetic training images. Real: Real training images. No real training images available for ITODD. 
{\dag} denotes the direct regression results of CosyPose (without using the pose refinement).
{*} denotes the results of SC6D without using the test-time augmentation. 
We highlight the best results in \textbf{bold} and \underline{underline} the second best results.
}
\label{tab:overall_eval}
\end{table*}

\subsection{Comparison to the state-of-the-art}

\paragraph{Quantitative evaluation.} 
The evaluation results are reported in Table \ref{tab:overall_eval} in terms of the AR metric. Overall, SC6D achieves comparable performance using neither the object 3D model nor prior knowledge of object symmetries. In particular, when trained using only synthetic PBR images, SC6D obtains 73.9\% AR on T-LESS, 30.3\% on ITODD and 61.1\% on YCB-V, and outperforms all baselines except SurfEmb \cite{haugaard2021surfemb}. After fine-tuning using the mixed synthetic and real training images, SC6D improves the performance to 78.0\% on T-LESS and 78.8\% on YCB-V and surpasses SurfEmb by 1.0\% and 7.0\%, respectively. It is worth noting that, given a single object crop, SC6D takes approximately 30ms for inference, while SurfEmb costs around 2200ms due to the time-consuming PnP-RANSAC \cite{fischler1981random} procedure. 

In addition, SC6D obtains substantially better results compared with another correspondence-free approach CosyPose \cite{labbe2020cosypose}. Specifically, CosyPose includes an initial pose regression network (CosyPose-coarse) and a pose refinement network (CosyPose-refiner). Without using the test-time augmentation, the plain SC6D outperforms the direct regression-based CosyPose-coarse by a large margin, \eg, 77.1\% \textit{vs.} 61.6\% on T-LESS, 29.5\% \textit{vs.} 13.1\% on ITODD, and 78.1\% \textit{vs.} 65.5\% on YCB-V. Furthermore, even equipped with the additional pose refinement network, CosyPose is still inferior to SC6D on all evaluations except on YCB-V with real images available.

\begin{table*}[t]
\small
\centering
\renewcommand{\arraystretch}{1.2}
\begin{tabular}{ c | c c c c c c | c c c | c c }
\hline
\shortstack{~\\ Row \\ ~} & \shortstack{ ~ \\ $\delta_z$ \\ Regression } & \shortstack{~ \\ $\delta_z$\\ Classification  } & \shortstack{~ \\ $\delta_z$ \\ Expectation} & \shortstack{~ \\ Focal \\ Loss } & \shortstack{~  \\ $\mathbf{M}^B_{kuv}$ \\ PE Map } & \shortstack{~\\ Test-time \\ Augmentation} & \shortstack{~\\ AR\textsubscript{VSD} \\ ~}  & \shortstack{~\\ AR\textsubscript{MSSD} \\ ~}  & \shortstack{~\\AR\textsubscript{MSPD} \\ ~}  & \shortstack{~\\ AR \\ ~}  \\ 
\hline
$A0$ & \checkmark & & & & &  & 0.552 & 0.598 & 0.825 & 0.658 \\ 
$A1$ & & \checkmark & & & & & 0.621 & 0.672 & 0.828 & 0.707 \\ 
$A2$ & & \checkmark & & \checkmark &  &  & 0.632 & 0.681 & 0.835 &  0.716 \\
$A3$ & & & \checkmark & \checkmark &  &  &  0.635 & 0.683 & 0.835 & 0.718 \\ 
$A4$ & & & \checkmark & \checkmark & \checkmark  &  & 0.649 & 0.700 & 0.839 & 0.729 \\ 
\hline
Ours & & & \checkmark & \checkmark & \checkmark &  \checkmark  & 0.662 & 0.712 &  0.844 & 0.739 \\ 
\hline
\end{tabular}
\caption{Ablation studies on \textbf{T-LESS} using only synthetic PBR images for training.
The 2D positional encoding (PE) map ( $\mathbf{M}^B_{kuv}$) is obtained by back-projecting the 2D pixel coordinates with the calibrated object-centric intrinsic $\mathbf{K}_B$.
The object location estimate along the z-axis can be obtained by either directly regressing the $\delta_z$, classifying the $\delta_z$ into the predefined labels, or computing the expected value of $\delta_z$ based on the estimated label probabilities.
}
\label{tab:ablate}
\end{table*}

\begin{figure*}[t]\centerline{\includegraphics[width=.95\linewidth]{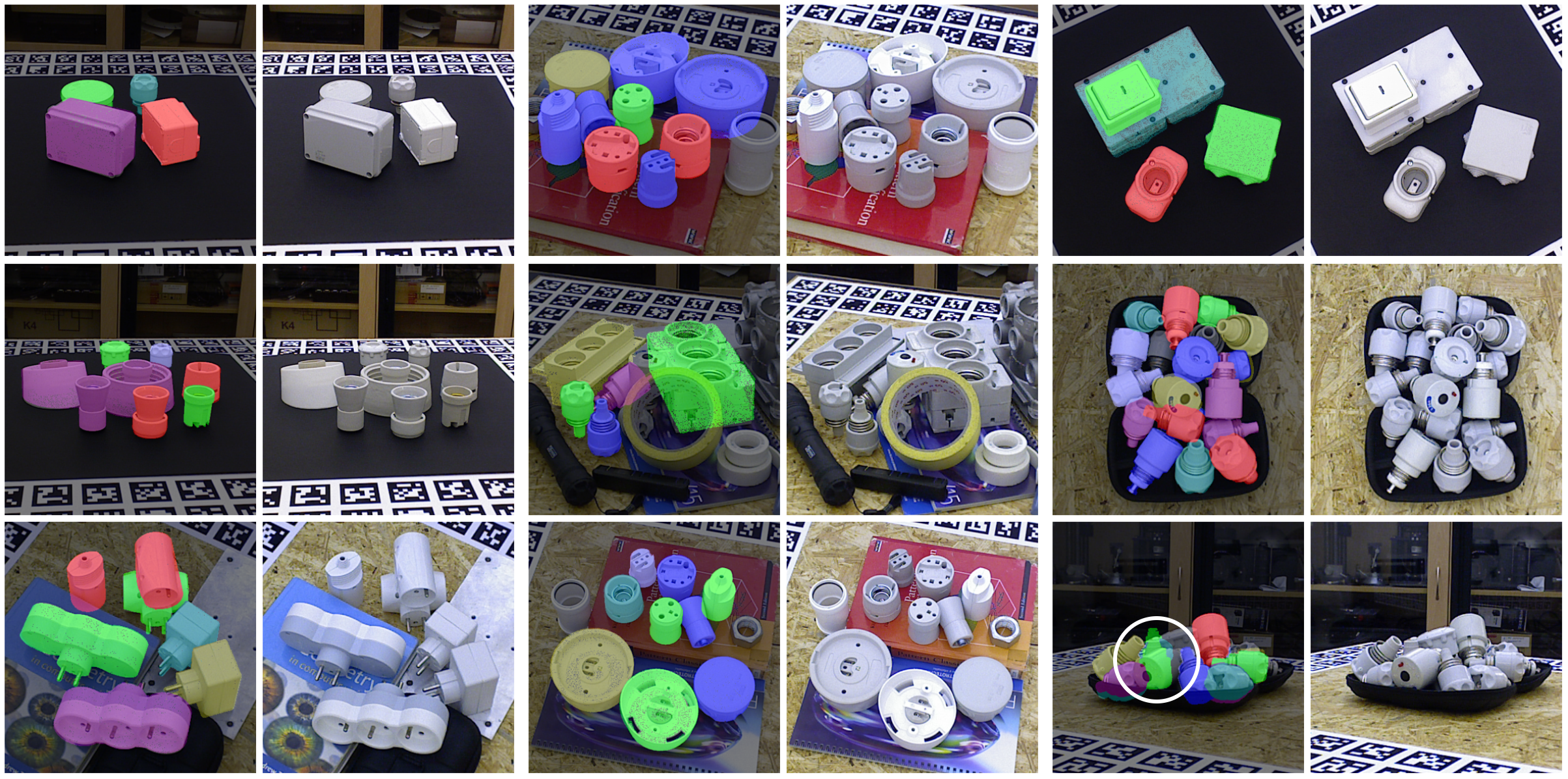}}
\caption{
Examples of qualitative evaluation on T-LESS. 
The object point clouds are transformed from the object coordinate system to the camera coordinate system using the estimated object 6D poses and overlaid on the RGB images with different colors. The original RGB images are shown on the right side of each group for reference. A failure case is presented on the right bottom and highlighted with a white circle.
} 
\label{fig:vis_tless}
\end{figure*}

\paragraph{Qualitative examples.}  Some examples for qualitative evaluation on T-LESS are shown in Figure \ref{fig:vis_tless}. We transform the object point clouds from the object coordinate system to the camera coordinate system using the estimated object 6D pose with the known camera intrinsic $\mathbf{K}_X$ and overlay the transformed point clouds on the RGB images.


\subsection{Ablation Studies}
We conduct extensive experiments on T-LESS to investigate the effect of each individual component of SC6D using only synthetic PBR \cite{hodan2020bop} images for training. The evaluation results are presented in Table \ref{tab:ablate}.

\vspace{-1mm}
\paragraph{Regression \textit{vs.} classification.} The classification-based object location  estimation ($\delta_z$) along the z-axis can obtain around 5\% improvement over the regression-based counterpart (row $A1$ \textit{vs.} row $A0$). This shows the effectiveness of the proposed classification based strategy. In addition, using the focal loss to train the $\delta_z$ classification task can improve the performance by around 1\% ($A2$ \textit{vs.} $A1$).
Further, a slightly better performance ($A3$ \textit{vs.} $A2$) can be obtained using the continuous z-axis translation estimates (\ie, the expectation value of $\delta_z$).

\vspace{-1mm}
\paragraph{2D positional encoding map.} Comparing $A4$ with $A3$, 1.1\% benefit can be obtained with the 2D positional encoding map ($\mathbf{M}^{B}_{kuv}$ PE Map) back-projected using the calibrated object-centric camera intrinsic $\mathbf{K}_B$. In particular, we can observe that the benefit (1.4\% for AR\textsubscript{VSD}, 1.7\% for AR\textsubscript{MSSD}, and 0.4\% for AR\textsubscript{MSPD}) mainly results from the (z-axis) distance-aware pose metrics, \ie, AR\textsubscript{VSD} and AR\textsubscript{MSSD}. We hypothesize that the calibrated 2D PE map $\mathbf{M}^{B}_{kuv}$ can provide helpful scale (distance) information consistent with the re-scaled object-centric crop $\mathbf{B}$.

\vspace{-1mm}
\paragraph{Runtime analysis.} On a desktop with an AMD 835 Ryzen 3970X CPU and an NVIDIA RTX3090 GPU, SC6D takes approximately 30ms (20ms without using the test-time augmentation) to infer the object 6D pose given an RGB image and an object bounding box.

\section{Discussion and Limitation}
We have evaluated SC6D on three benchmark datasets, especially on T-LESS, regarding object ambiguities, like symmetries caused by poor texture, occlusion, etc. SC6D achieves an impressive trade-off between efficiency and accuracy. However, SC6D predicts the object 6D pose estimation from the global object appearance presented in an RGB image. As a holistic approach, SC6D may be inherently inferior to the correspondence-based methods in handling heavy occlusion, especially for ambiguity-free objects. In our future work, we attempt to explore the possibility of performing dense pose estimation based on the local appearance of the target object to mitigate this problem.

\section{Conclusion}
This work presents a symmetry-agnostic framework named SC6D for object 6D pose estimation from a single RGB image. An SO(3) encoder is proposed to encode the 3D rotation samples from SO(3) space and account for the object 3D orientation estimation. Thanks to the proposed SO(3) encoder, neither the prior knowledge of symmetries nor the 3D CAD model of an object is required for training and inference. In addition, we revisited the transformation from the original RGB image coordinate system to the object-centric crop coordinate system, based on which we can further improve the accuracy of the object location estimation along the z-axis. 

\section{Acknowledgement}
This work was supported by the Academy of Finland under the project \#327910.

{\small
\bibliographystyle{ieee_fullname}
\bibliography{egbib}
}

\newpage
\appendix
\section{Supplementary Materials}

\subsection{SO(3) Sampling}
At inference, we uniformly sample 4000 viewpoints on a sphere (following \cite{cai2022ove6d}) and 120 in-plane rotations for each viewpoint, which results $480000=120 \times 4000$ rotation samples. Table \ref{table:so3} presents the test-time trade-offs between efficiency and performance on T-LESS in terms of a varying number of SO(3) samplings. Intuitively, finer sampling results in better performance but higher runtime. We consider 480k SO(3) samplings as a good trade-off for inference in our experiments. In contrast, a coarse online sampling strategy (\eg, 5000 rotation samplings) is adopted  during the training time for speeding up the training process.

\begin{table}[h]
\small
\centering
\renewcommand{\arraystretch}{1.2}
\begin{tabular}[0.9\textwidth]{ c | c c c c c} 
\hline
\shortstack{ SO(3) samples} & \shortstack{20k} & \shortstack{60k} & \shortstack{180k} & \shortstack{480k} & \shortstack{1440k} \\
\hline
AR & 0.725 & 0.732 & 0.737 & 0.739 & 0.741 \\
\hline
Runtime (ms/obj) & 24 & 25 & 27 & 30 & 44 \\
\end{tabular}
\vspace{1pt} 
\caption{Evaluation on T-LESS in terms of a varying number of SO(3) sampling.}
\label{table:so3}
\end{table}

\subsection{SO(3) Encoder \textit{vs.} Implicit-PDF}
We further replace the SO(3) Encoder module in SC6D with Implicit-PDF \cite{murphy2021implicit}, \ie, the image embedding vector (extracted from the Pose Decoder) is first concatenated with a query SO(3) sample and then is utilized to estimate the log probability of the query. As a result, the variant with Implicit-PDF as the SO(3) prediction module obtains 0.720 AR on T-LESS, which is inferior to SC6D with SO(3) Encoder (0.73.9 AR). The reason could be that the framework of SC6D is specifically designed for the proposed SO(3) Encoder but is not optimal for Implicit-PDF.

\subsection{Architecture Details}
\begin{figure*}[h]
\centerline{\includegraphics[width=0.99\linewidth]{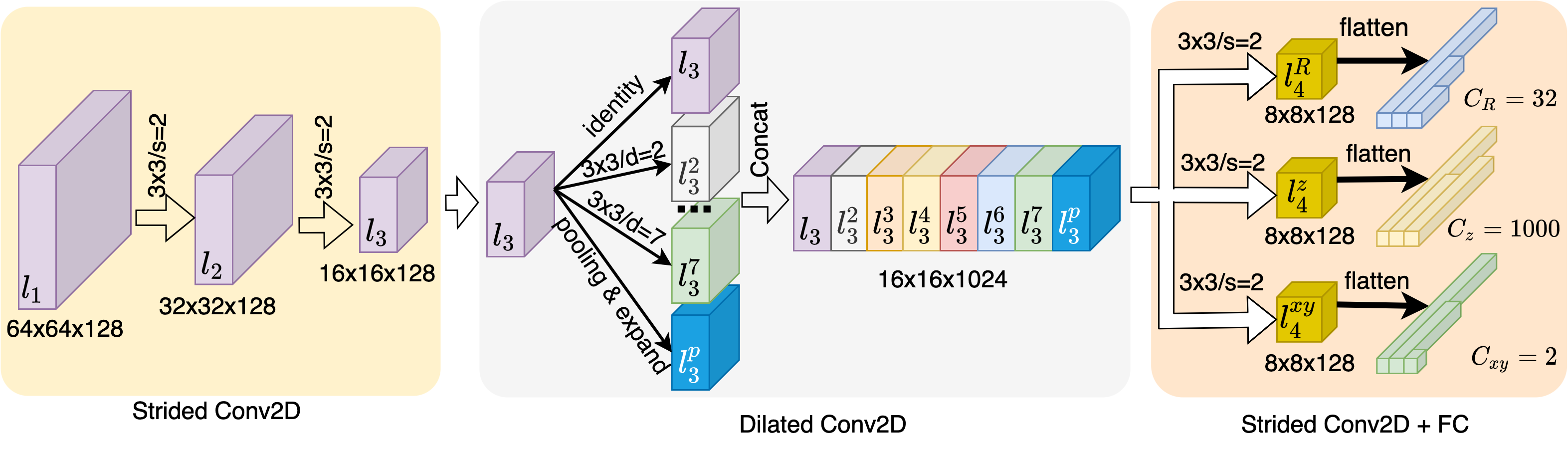}}
\caption{
Structure details of the pose decoder. The pose decoder consists of three subsequent sub-modules. The input features ($l_1$) are first down-sampled by two subsequent $3 \times 3$ Conv2D layers with the stride $s=2$. In the second sub-module, the down-sampled features ($l_3$) are processed by 6 parallel $3 \times 3$ Conv2D layers with the dilations $d=\{2,3,4,5,6,7\}$ and by a global average pooling (and expanding) to aggregate the information at different scales . We concatenate these 7 feature maps (each with the size $16\times 16\times 128$) to the down-sampled one ($l_3$) and feed the concatenated feature maps to three parallel decoding heads. The decoding heads share a similar structure (one $3\times 3$ Conv2D with the stride $s=2$ followed by two FC layers) but with different output dimensions ($C_R=32, C_{xy}=2, C_z=1000$ in our experiments). Note that each layer is followed by a GroupNorm and ReLU layers except the final output layer.
} 
\label{fig:pose_dec}
\end{figure*}

The architecture details of the pose decoder is illustrated in the Figure \ref{fig:pose_dec}.

\subsection{Qualitative Evaluation on YCB-V}
\begin{figure*}[h]\centerline{\includegraphics[width=0.99\linewidth]{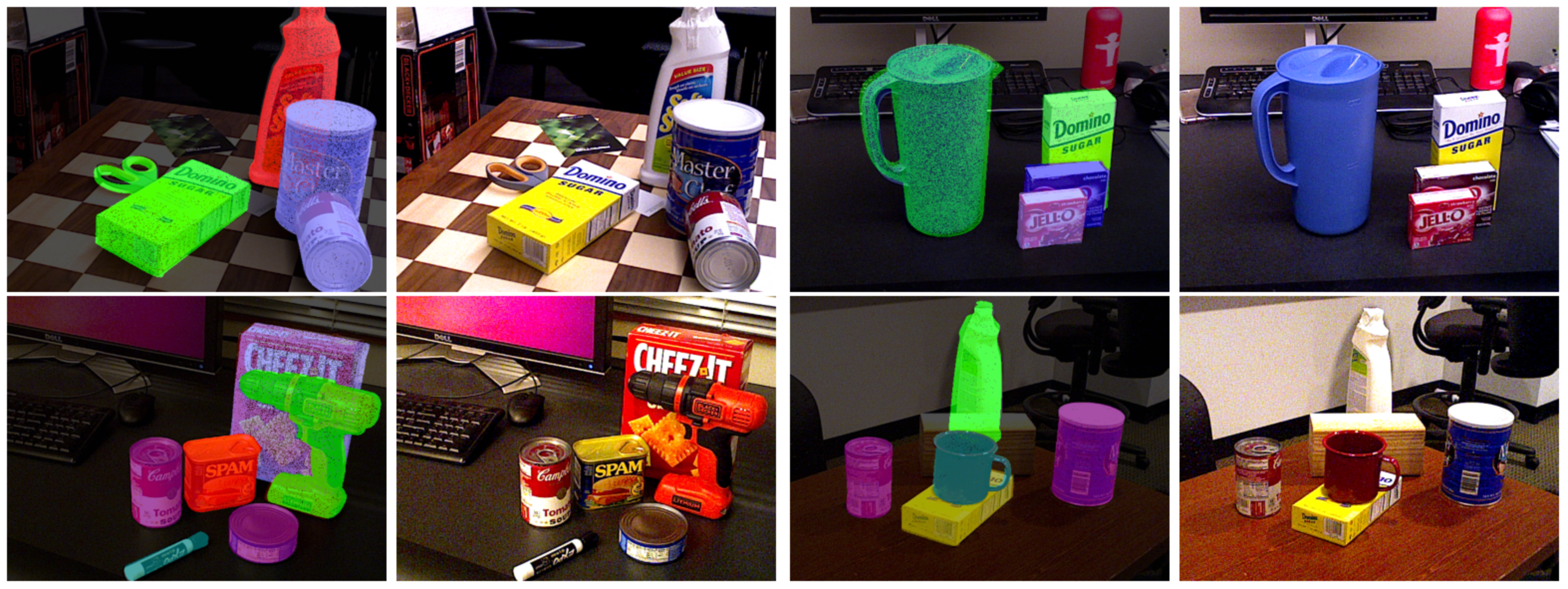}}
\caption{
Examples of qualitative evaluation on YCB-V. 
The object point clouds are transformed from the object coordinate system to the camera coordinate system using the estimated object 6D poses and overlaid on the RGB images with different colors. The original RGB images are shown on the right side of each group for reference. 
} 
\label{fig:vis_ycbv}
\end{figure*}

Some examples for qualitative evaluation on YCB-V are shown in Figure \ref{fig:vis_ycbv}. We transform the object point clouds from the object coordinate system to the camera coordinate system using the estimated object 6D pose with the known camera intrinsic $\mathbf{K}_X$ and overlay the transformed point clouds on the RGB images with different colors.

\end{document}